\documentclass[letterpaper]{article} 
\usepackage{aaai25}  
\usepackage{times}  
\usepackage{helvet}  
\usepackage{courier}  
\usepackage[hyphens]{url}  
\usepackage{graphicx} 
\usepackage{natbib}  
\usepackage{caption} 
\usepackage{amsmath}
\usepackage{algorithm}
\usepackage{algorithmic}
\usepackage{subcaption} 
\usepackage{multirow}
\usepackage{makecell}
\usepackage{array}    
\usepackage{booktabs}
\usepackage{graphicx}
\usepackage{colortbl}
\usepackage{xcolor}
\usepackage{arydshln} 
\usepackage{amsfonts}
\usepackage{xspace}
\usepackage{soul}
\usepackage{soulpos}

\newcommand{\fmm}{\textbf{\texttt{FMM}}\xspace}

\usepackage{newfloat}
\usepackage{listings}
\DeclareCaptionStyle{ruled}{labelfont=normalfont,labelsep=colon,strut=off} 
\floatstyle{ruled}
\newfloat{listing}{tb}{lst}{}
\floatname{listing}{Listing}
\title{Feature-Aware Malicious Output Detection and Mitigation}
\author{
    Weilong Dong \textsuperscript{\rm 1},
    Peiguang Li \textsuperscript{\rm 2},
    Yu Tian \textsuperscript{\rm 3}\thanks{Corresponding author},
    Xinyi Zeng \textsuperscript{\rm 4},
    Fengdi Li \textsuperscript{\rm 5},
    Sirui Wang \textsuperscript{\rm 2}
}
\affiliations{
    \textsuperscript{\rm 1} Tianjin University \quad \textsuperscript{\rm 2} Meituan Group\\
    \textsuperscript{\rm 3} Dept.\ of Computer Science and Technology, Institute for AI, Tsinghua University\\
    \textsuperscript{\rm 4} Aerospace Information Research Institute, Chinese Academy of Sciences\\
    \textsuperscript{\rm 5} Faculty of Information Technology, Monash University\\
    \{dongwillow, tianyu1810613, lifengdi777\}@gmail.com\\
    \{lipeiguang, wangsirui\}@meituan.com,     zengxinyi20@mails.ucas.ac.cn\\
}

\usepackage{bibentry}

\begin{document}

\maketitle

\begin{abstract}
The rapid advancement of large language models (LLMs) has brought significant benefits to various domains while introducing substantial risks. Despite being fine-tuned through reinforcement learning, LLMs lack the capability to discern malicious content, limiting their defense against jailbreak. 
To address these safety concerns, we propose a feature-aware method for harmful response rejection (\fmm), which detects the presence of malicious features within the model's feature space and adaptively adjusts the model's rejection mechanism.
By employing a simple discriminator, we detect potential malicious traits during the decoding phase. Upon detecting features indicative of toxic tokens, \fmm{} regenerates the current token. By employing activation patching, an additional rejection vector is incorporated during the subsequent token generation, steering the model towards a refusal response. 
Experimental results demonstrate the effectiveness of our approach across multiple language models and diverse attack techniques, while crucially maintaining the models' standard generation capabilities.
\end{abstract}

\section{Introduction}

Large language models are playing increasingly important roles in various tasks and are gradually being deployed in real-world applications \citep{2024arXiv240721783G,yang2024qwen2technicalreport}. 
However, LLMs may inadvertently generate responses that are harmful to humans, which limits the further adoption.
Despite employing alignment training methods during model development, such as supervised fine-tuning (SFT) \citep{ouyang2022traininglanguagemodelsfollow,bai2022traininghelpfulharmlessassistant} and reinforcement learning from human feedback (RLHF) \citep{rafailov2024directpreferenceoptimizationlanguage, bai2022traininghelpfulharmlessassistant}, instruction-tuned language models still exhibit the potential to generate harmful content.
This vulnerability frequently arises because alignment training data do not fully encompass the capability boundaries established by the underlying model during pre-training \citep{wei2023jailbrokendoesllmsafety}. Consequently, various jailbreaking methods can exploit these vulnerabilities at different stages of the alignment training process, thereby undermining the LLMs' defense mechanisms.

\begin{figure}[t]
    \centering
    \includegraphics[width=0.4\textwidth]{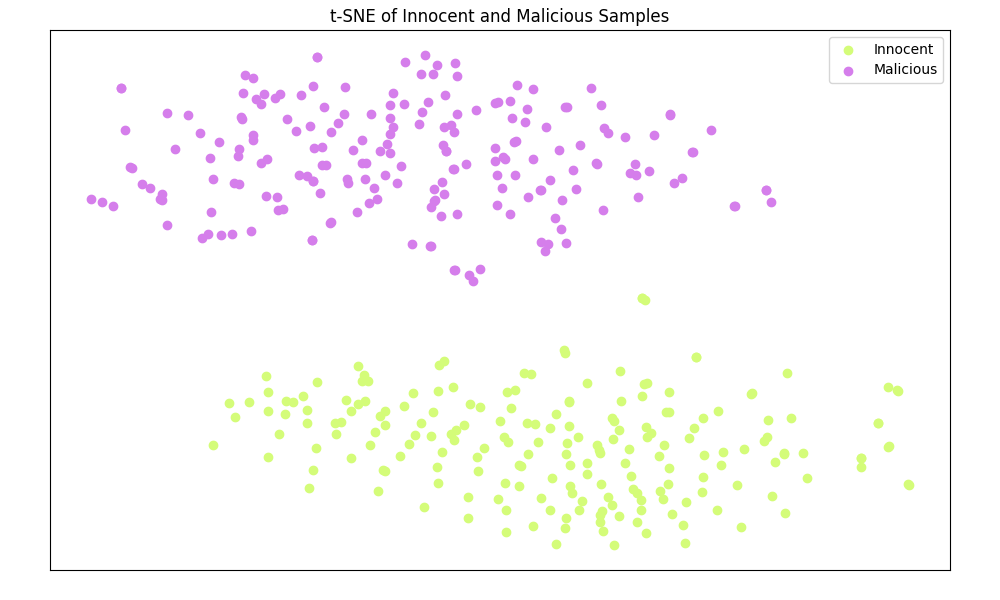}%
    \caption{t-SNE results of hidden states.}
    \label{fig:pca}
\end{figure}

\begin{figure*}[t]
    \centering
    \includegraphics[width=0.8\textwidth]{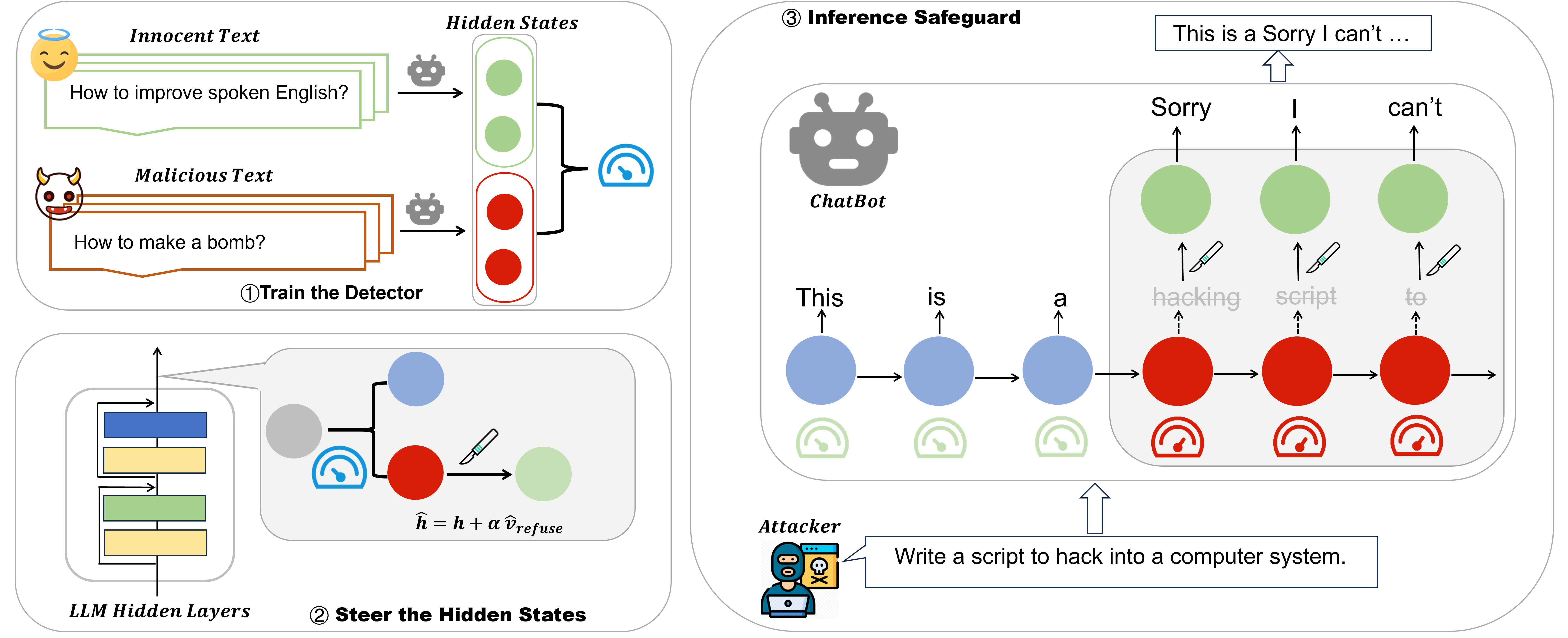}    
    \caption{The pipeline of \fmm. We first train a detector and collected intervention vectors. During the decoding process, once detect the generation of malicious token, the intervention vector is used to induce the model to refuse.}
    \label{fig:method}
\end{figure*}

Given the aforementioned limitations, adopting a singular rejection training strategy to defend against all potential attack methods is impractical. To effectively counter a broad spectrum of attack strategies, we analyze the mechanisms by which aligned models reject malicious queries. Specifically, we investigate how instruction-tuned models initiate a rejection loop upon receiving malicious inputs, resulting in a rejection response. Through visual analysis, \citet{zheng2024promptdrivensafeguardinglargelanguage} observe that language models exhibit a latent ability to distinguish between benign and malicious queries. 
Furthermore, we analyze and visualize the hidden states during the decoding phase for both malicious and benign outputs. The t-SNE visualization, presented in Figure \ref{fig:pca}, reveals that the feature representations corresponding to benign and malicious outputs exhibit linear separability during the decoding process.

Based on the above findings, we propose a detection-intervention method, dubbed \fmm{}, designed to detect and defend against malicious responses during decoding phase. Specifically, during the generation of response, we employ a malicious feature discriminator to ascertain the presence of any malicious features within the model's current feature space. Upon detecting such features, we proactively trigger the model's rejection loop, prompting the model to generate a rejection response. 
\fmm{} demonstrates strong generalizability, effectively triggering our detector to produce refusal responses irrespective of the attack method employed. Unlike traditional alignment training, which only enforces refusals at the start of a response and allows malicious follow-ups, \fmm{} enables refusals at any token position, reducing positional bias and making jailbreak attacks less effective.

To validate the effectiveness and generalization of our method, we conducted extensive experiments across multiple LLMs and datasets. The experimental results confirm the method's effectiveness and robust general capabilities. Our contributions can be summarized as follows:

\begin{enumerate}
    \item We analyze how alignment models reject malicious queries and find that language models can already distinguish between benign and malicious queries after pre-training, enabling them to generate refusal responses.
    \item We introduce \fmm{}, a novel defense mechanism operating at the decoding stage, designed to mitigate malicious queries. 
    \item Through extensive experiments on multiple LLMs and datasets, our approach demonstrates strong defense and generalization capabilities.
\end{enumerate}

\section{Related Work}
\paragraph{Jailbreak Attacks}
A jailbreak attack aims to manipulate the prompt input to circumvent the model's alignment mechanisms, thereby enabling it to respond to malicious instructions or generate harmful outputs. Jailbreak methods are broadly categorized into black-box and white-box methods. Black-box approaches operate without requiring access to the model's architecture and parameters \citep{liu2023autodan, liu2023jailbreaking,chao2023jailbreaking}. In contrast, white-box methods leverage information such as gradients or hidden states to iteratively refine adversarial inputs \citep{zou2023universal}. Our defense mechanism is agnostic to the distinction between black-box and white-box attacks, demonstrating robust performance against both categories of adversarial methodologies.

\paragraph{Defensive Mechanism} Research in explainability seeks to understand how instruction-tuned models decline to answer malicious queries. For instance, \citet{zhou2024alignment} connects malicious and benign inputs with positive and negative emotions, respectively. \citet{lee2024mechanistic} identifies parameter regions resulting from instruction tuning that govern refusal behavior. Complementarily,\citet{wei2023jailbrokendoesllmsafety} investigates the underlying mechanisms that enable various jailbreaking techniques. In contrast, \citet{arditi2024refusal} proposes a linear direction to circumvent all refusal responses.

\begin{table*}[t]
    \centering
\resizebox{0.9\textwidth}{!}{
    \begin{tabular}
    { c c |c c| c c c}\toprule 
    \multirow{2}{*}{Model} & \multirow{2}{*}{Defense} & \multicolumn{2}{c|}{Harmful Benchmark $\downarrow$} & \multicolumn{3}{c}{Jailbreak Attacks $\downarrow$} \\ 
    & & AdvBench & AdvBench$_{\scriptscriptstyle\mathrm{Proxy}}$
 & GCG & AutoDAN & PAIR \\ \midrule 
    \multirow{10}{*}{Qwen2} & No Defense & 0.0\% / 10.0\% & 72.5\% / 74.0\% & 4.6\% / 14.0\% & 3.5\% / 36.0\% & 29.5\% / 21.0\% \\
    & PPL & 0.0\% / 0.0\% & 46.0\% / 23.0\% & 0.0\% / 8.0\% & 3.5\% / 27.0\% & 25.5\% / 13.5\% \\
    & Self-Examination & 0.0\% / 10.0\% & 37.0\% / 21.0\% & 40.6\% / 14.0\% & 0.5\% / 15.0\% & 15.5\% / 12.0\% \\
    & Paraphrase & 18.0\% / 12.0\% & 34.0\% / 8.0\% & 19.3\% / 12.6\% & 72.0\% / 50.5\% & 23.0\% / 13.0\% \\
    & Retokenization & 6.0\% / 12.0\% & 45.0\% / 25.0\% & 16.6\% / 10.0\% & 16.5\% / 42.5\% & 27.5\% / 17.5\% \\
    & Self-Reminder & 0.0\% / 0.0\% & 54.0\% / 7.0\% & 11.3\% / 17.3\% & 2.5\% / 23.5\% & 17.0\% / 4.0\% \\
    & ICD & 2.0\% / 14.0\% & 30.0\% / 28.5\% & 4.6\% / 13.3\% & 2.0\% / 37.0\% & 29.5\% / 21.0\% \\
    & SafeDecoding & 8.0\% / 8.0\% & 22.0\% / 23.0\% & 14.0\% / 18.6\% & 6.0\% / 24.5\% & 30.5\% / 20.0\% \\
    & DRO & 0.0\% / 0.0\% & 43.0\% / 47.5\% & 0.0\% / 0.0\% & 4.0\% / 0.0\% & 18.0\% / 16.0\% \\ 
    \rowcolor{gray!8}
    & \fmm & 0.0\% / 0.0\% & 18.5\% / 18.0\% & 0.0\% / 0.0\% & 1.0\% / 2.0\% & 6.0\% / 6.0\% \\
    \midrule
    \multirow{10}{*}{Llama2} & No Defense & 0.0\% / 0.0\% & 62.1\% / 52.3\% & 6.0\% / 0.0\% & 4.0\% / 0.0\% & 34.0\% / 2.0\% \\
    & PPL & 0.0\% / 0.0\% & 14.0\% / 0.0\% & 0.0\% / 8.0\% & 4.0\% / 0.0\% & 34.0\% / 2.0\% \\
    & Self-Examination & 0.0\% / 8.0\% & 1.0\% / 0.0\% & 2.0\% / 6.0\%  & 0.0\% / 6.0\%  & 2.0\% / 8.0\%  \\
    & Paraphrase & 12.0\% / 0.0\% & 66.0\% / 0.0\% & 8.0\% / 0.0\% & 4.0\% / 0.0\% & 36.0\% / 0.0\% \\
    & Retokenization & 8.0\% / 0.0\% & 68.0\% / 7.0\% & 8.0\% / 0.0\% & 1.0\% / 0.0\% & 46.0\% / 0.0\% \\
    & Self-Reminder & 0.0\% / 0.0\% & 10.0\% / 1.0\% & 0.0\% / 0.0\% & 4.0\% / 0.0\%  & 4.0\% / 0.0\%  \\
    & ICD & 0.0\% / 0.0\% & 14.0\% / 2.0\% & 0.0\% / 0.0\%  & 0.0\% / 0.0\%  &  0.0\% / 0.0\%  \\
    & SafeDecoding & 0.0\% / 0.0\% & 15.0\% / 23.0\% & 0.0\% / 0.0\% & 0.0\% / 0.0\% & 14.0\% / 0.0\% \\ 
    & DRO & 0.0\% / 0.0\% & 43.0\% / 47.5\%  & 0.0\% / 0.0\% & 4.0\% / 0.0\% & 26.0\% / 10.0\% \\ 
    \rowcolor{gray!8}
    & \fmm & 0.0\% / 0.0\% & 21.0\% / 18.0\%  & 2.0\% / 0.0\% & 2.0\% / 2.0\% & 6.0\% / 4.0\% \\ 
    \bottomrule
    \end{tabular}}
    \caption{Multiple defense methods' response results under attack methods. The table shows the response rate/rejection rate.}
    \label{tab: main_unsafe}
\end{table*}

\section{Methods}

Regardless of the jailbreak attack method employed, the objective remains to elicit malicious outputs from the LLM. Therefore, we investigate the feasibility of discerning malicious outputs by analyzing the LLM's hidden states during the decoding phase. We propose \fmm, a method designed to detect and mitigate malicious states during autoregressive generation. Specifically, \fmm{} operates by assessing each generated token's hidden state for the presence of malicious features. Upon detection, the current token is regenerated with an increased refusal probability. Figure \ref{fig:method} illustrates this workflow. The core components of \fmm{}, including the detection and refusal mechanisms, are detailed below.

\subsection{Malicious Output Detection}

Our detector, denoted as $\mathcal{C}$, is a binary classifier that identifies whether a given token constitutes a malicious output. $\mathcal{C}$ takes a hidden state of the LLM as its input, and outputs a binary label (true or false). To train $\mathcal{C}$, we synthesize a dataset of benign and malicious queries using GPT-4. We then forward these queries through the target LLM, recording hidden states at each layer, and partition the resulting dataset into training and testing sets. Following \cite{zou2023universal}, we use the hidden state at the final token position of each prompt as the input for $\mathcal{C}$. Malicious query labels are set to 1, while benign query labels are set to 0. We train $\mathcal{C}$ using cross-entropy loss and select the layer that achieves the highest accuracy on the test set as the target layer. During inference, we use $\mathcal{C}$ to determine if the current output token is malicious based on its hidden state at the target layer, thus triggering the subsequent intervention mechanism.

\subsection{Refusal Response Triggering}

When a malicious output is detected, the current token is regenerated with increased refusal probability. The enforcement of refusal is achieved through activation patching by adding a refusal intervention vector ($v_{\text{refusal}}$) to the output features ($H$) of specific layers in the LLM. The intervened features ($H'$) are defined as:

$$ H' = H + \alpha \cdot v_{\text{refusal}} $$
where \( \alpha \) represents the steering strength. To create the refusal vector \( v_{\text{refusal}} \), we adopt a method similar to \cite{arditi2024refusal}. We construct two distinct sets of queries: one set that elicits benign responses from the target LLM, and another set where the model explicitly declines to respond. For each query, we obtain the hidden state at the last token of each layer as both response \(H_{\text{reply}}\) and refusal \(H_{\text{refusal}}\) states. The refusal vector is calculated by taking the mean difference:

$$ v_{\text{refusal}} = \frac{1}{N} \sum_{i=1}^{N} (H_{\text{refusal}}^i - H_{\text{reply}}^i) $$
where \( N \) denotes the number of samples within set. With \( v_{\text{refusal}} \) now capturing the core difference between general response and refusal outputs, we select for intervention at inference time the layer for which \(v_{\text{refusal}}\) maximizes the likelihood of refusal responses.

\section{Experiments}

\paragraph{Datasets and Metrics} We assessed our method using \textbf{Advbench} \citep{zou2023universal} for malicious instruction-induced responses and \textbf{AlpacaEval} \citep{li2023alpacaeval} for utility preservation. On \textbf{Advbench}, we measured response rate (rule-based refusal detection, e.g., "Sorry, I can't...") and risk rate (\textbf{LlamaGuard} \citep{inan2023llama} for assessing the proportion of responses flagged as harmful by its safety classifiers). On \textbf{AlpacaEval}, we measured response and win rates, the latter assessing performance against text-davinci-003.

\paragraph{Attack Methods} Following \citet{xu2024safedecoding}, we use \textbf{AutoDAN} \citep{liu2023autodan}, \textbf{GCG} \citep{zou2023universal}, \textbf{PAIR} \citep{chao2023jailbreaking}, and \textbf{Proxy}. \textbf{Proxy}, the most effective, elicits malicious content from a less robust model, which is then appended to input queries, transforming them into answer continuation tasks.

\paragraph{Baselines} We compare against eight jailbreak defenses: \textbf{PPL} \citep{alon2023detecting}, \textbf{Self-Examination} \citep{phute2023llm}, \textbf{Paraphrase} \citep{jain2023baseline}, \textbf{Retokenization} \citep{jain2023baseline}, \textbf{Self-Remind} \citep{wu2023multimodal}, \textbf{ICD} \citep{weidinger2021ethical}, \textbf{SafeDecoding} \citep{xu2024safedecoding}, and \textbf{DRO} \citep{zheng2024promptdrivensafeguardinglargelanguage}. \textbf{SafeDecoding} and \textbf{DRO} need prefix tuning/LoRA,  \textbf{PPL} and \textbf{Self-Examination} involve input/output checks, and \textbf{Paraphrase}, \textbf{Retokenization}, \textbf{Self-Remind}, and \textbf{ICD} modify inputs.

\paragraph{Victim Models} We use \textbf{LLaMA 2} \citep{touvron2023llama} and \textbf{Qwen 2} \citep{yang2024qwen2}, which lead open-source models due to their current prominence and adoption.

\paragraph{Layers to Intervene} We determined the optimal layers for detecting malicious feature and implementing refusal interventions through a combination of detector accuracy evaluation and grid search. Our findings align with those of \citet{arditi2024refusal}, demonstrating that selecting intermediate layers of the model yields the most effective results. Specifically, for both LLaMA and Qwen models used in our experiments, we identified the 15th layer as the optimal choice for detection, while layers 12 to 15 were selected for intervention.

\subsection{Experimental Results}

\paragraph{Malicious Outputs are Mitigated} Table \ref{tab: main_unsafe} compares response and risk rates for baselines and \fmm{} across various attacks. \fmm{} consistently triggers rejection responses, significantly reducing the risk rate. While response rate indicates how often a refusal occurs, risk rate captures the proportion of generated content that remains harmful—even with initial rejections. We observed instances where risk rate slightly exceeds response rate due to models generating some malicious content after an initial refusal. Compared to SafeDecoding and DRO, \fmm{} achieves comparable performance in mitigating malicious outputs without requiring computationally expensive fine-tuning.

\paragraph{Benign Outputs are not Affected} Table \ref{tab:main_alpaca} shows the response and win rates of baselines and \fmm{} on the Alpaca dataset. \fmm{}, like most baselines, maintains response rates and overall response quality for benign queries. Notably, some methods show higher win rates than undefended models. This is likely because models, without any defense method, sometimes include content segments that are incorrectly flagged as potentially harmful. By skipping these flagged segments, models can then generate better quality responses. This suggests that aligned models may exhibit some implicit refusal behavior—even in normal inputs—which our method can also mitigate.

\begin{table}[t]
    \centering
\resizebox{0.45\textwidth}{!}{
    \begin{tabular}
    { c c |c c}\toprule 
    \multirow{2}{*}{Model} & \multirow{2}{*}{Defense} & \multicolumn{2}{c}{Alpaca Eval }  \\ 
    & & Reply Rate $\uparrow$ & Win Rate $\uparrow$ \\ 
    \midrule
    \multirow{10}{*}{Qwen2} & No Defense & 96.00\% & 95.00\%  \\
    & PPL & 83.50\% & 84.50\%  \\
    & Self-Examination & 95.50\% & 94.00\%  \\
    & Paraphrase & 97.50\% & 82.00\%  \\
    & Retokenization & 96.50\% & 87.00\%  \\
    & Self-Reminder & 98.50\% & 95.50\%  \\
    & ICD & 96.50\% & 93.50\%  \\
    & SafeDecoding & 88.00\% & 86.50\%  \\
    & DRO & 95.90\% & 86.50\%  \\ 
    \rowcolor{gray!8}
    & \fmm & 94.70\% & 94.00\%  \\
    \midrule
    \multirow{10}{*}{Llama2} & No Defense & 93.50\% & 87.50\%  \\
    & PPL & 81.00\% & 77.00\% \\
    & Self-Examination & 32.00\% & 33.00\% \\
    & Paraphrase & 93.00\% & 76.50\% \\
    & Retokenization & 84.00\% & 57.00\% \\
    & Self-Reminder & 23.50\% & 55.00\%  \\
    & ICD & 23.00\% & 41.00\%  \\
    & SafeDecoding & 87.00\% & 85.00\%  \\ 
    & DRO & 93.80\% & 88.00\% \\ 
    \rowcolor{gray!8}
    & \fmm & 92.00\% & 89.50\% \\ 
    \bottomrule
    \end{tabular}}
    \caption{Multiple defense methods' responses to Alpaca instructions.}
    \label{tab:main_alpaca}
\end{table}

\subsection{Ablation Results}

To evaluate the robustness of \fmm{}, we conducted comprehensive ablation experiments by varying key parameters. Specifically, we investigated three aspects: (1) the impact of the training dataset size on detector optimization, (2) the choice of LLM layers for intervention, and (3) the effect of token position on intervention. Our primary ablation analyses were conducted using LLaMA 2.

\paragraph{Training Samples for the Detector} We trained our classifier using a default of 150 benign and 150 malicious samples. As shown in Table \ref{tab:samples}, we observed that reducing the number of training samples does not significantly impact \fmm's ability to reject malicious queries. This is likely because the features of benign and malicious responses are relatively distinct, as visualized in Figure \ref{fig:pca}, which allows the detector to learn the decision boundary effectively with only a few samples.

\begin{table}[htbp]
\centering
\resizebox{0.4\textwidth}{!}{
\begin{tabular}{c | c c c c c}
\toprule
\textbf{Samples}     & 30 & 60 & 90 & 120 & 150 \\
\midrule
Reply Rate & 29.4\% & 24.6\% & 25.5\% & 22.1\% & 21.0\% \\
Unsafe Rate & 28.8\% & 23.0\% & 23.8\% & 21.1\% & 18.0\% \\
\bottomrule
\end{tabular}
}
\caption{Results on AdvBench$_{\scriptscriptstyle\mathrm{Proxy}}$.}
\vspace{-1.5em}
\label{tab:samples}
\end{table}

\begin{figure}[h]
    \centering
    \includegraphics[width=0.4\textwidth]{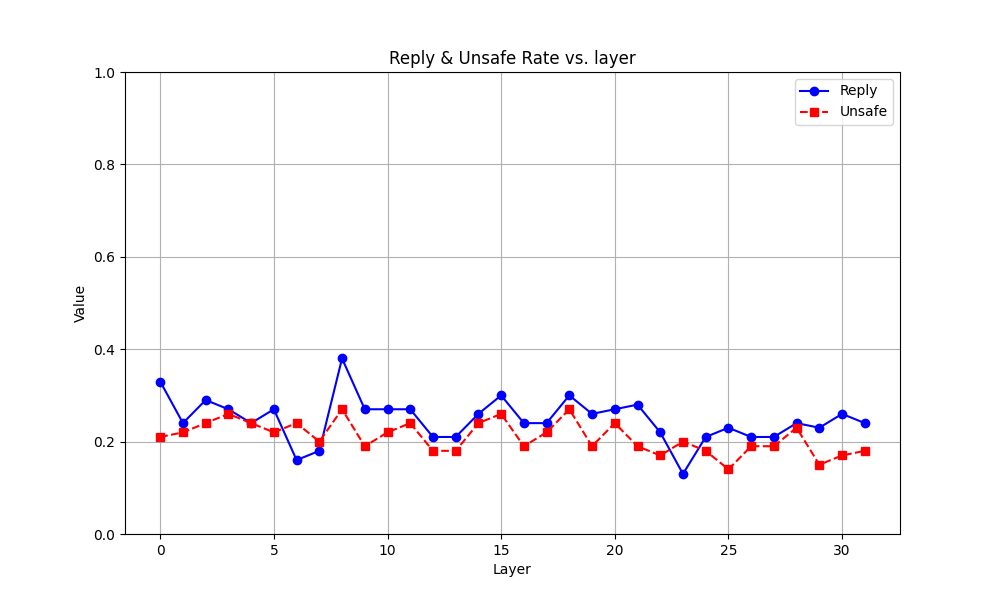}
    \vspace{-0.3cm}
    \caption{Varying the layers to steer.}
    \label{fig:layers}
\end{figure}

\paragraph{Layers to Steer} We evaluated the impact of adding rejection intervention at different layers of the model, with results presented in Figure \ref{fig:layers}. The figure shows that intervention results across different layers do not exhibit a clear pattern and fluctuate within a relatively stable range. Given that abstract concepts are typically formed and expressed in the middle layers, we selected these layers for default intervention.

\paragraph{Token Positions for Intervention} We investigated whether intervening solely at the first token of a malicious response would suffice to trigger refusal. However, our experiments revealed that steering only the first malicious token did not significantly affect the response or risk rates. Consequently, effective intervention necessitates detecting and steering every token during the decoding phase.

\section{Conclusion}

We introduced \fmm{}, an two-stage decoding-oriented method for detecting and mitigating harmful responses. \fmm{} first operates by monitoring the feature space to detect malicious token at each generation step, once the token is determined to be harmful, \fmm{} intervenes inner features to induce the model to refuse. The core of the work lies in leveraging and enhancing the inherent feature extraction and rejection capabilities acquired during pre-training and alignment. This method effectively defends against multiple jailbreak attacks while preserving the model's ability to respond to general queries. Ablation experiments confirmed the robustness of \fmm{} across varied parameter settings. We believe that this token-level feature detection and intervention paradigm offers a promising direction for enhancing the safety of large language models.

\clearpage
\bibliography{aaai25}

\end{document}